\documentclass[journal]{IEEEtran}
\usepackage{amsmath,amsfonts}
\usepackage{algorithmic}
\usepackage{algorithm}
\usepackage{array}
\usepackage[caption=false,font=normalsize,labelfont=sf,textfont=sf]{subfig}
\usepackage{textcomp}
\usepackage{stfloats}
\usepackage{url}
\usepackage{verbatim}
\usepackage{graphicx}
\usepackage{cite}
\begin{document}

\title{Physics-Based Deep Spatiotemporal Hyperlocal Radar Nowcasting with a Multi-Variable U-Net for High-Resolution Precipitation Forecasting}

\author{Akshay Sunil, Muhammed Rashid, Raja Sekhar Sivaraju, Sushma Nair, and Subimal Ghosh
\thanks{Akshay Sunil, Subimal Ghosh, and Muhammed Rashid are with the Centre for Climate Studies, Indian Institute of Technology Bombay, Mumbai, India (e-mail:
akshaysunil172@gmail.com; subimal@iitb.ac.in; muhammedrashid.edu@gmail.com).}
\thanks{Raja Sekhar Sivaraju and Sushma Nair are with the Regional Meteorological Centre (RMC) in Colaba, Mumbai, India (e-mail: sivaraju.rajasekhar@gmail.com;sushma.nair@imd.gov.in).}}

\markboth{}%
{Sunil \MakeLowercase{\textit{et al.}}: Physics-Based Deep Spatiotemporal Hyperlocal Radar Nowcasting with a Multi-Variable U-Net
for High-Resolution Precipitation Forecasting}

\maketitle

\begin{abstract}
Precipitation nowcasting over the immediate 10-
90 min period is important for flood management and real-
time decision-making in urban regions. Conventional short-range
forecasting with high-resolution numerical weather 
prediction requires frequent data assimilation, model initialization, and
spin-up, introducing computational latency that limits rapid operational guidance. 
Machine learning provides an alternative 
by learning storm evolution directly from high-frequency 
observations and producing forecasts quickly after training. 
This is particularly relevant for Mumbai, India, where monsoon convection, 
land-sea interactions, and localized intense rainfall make short-term prediction 
difficult. Here, we develop a compact radar-only nowcasting framework that 
combines multi-elevation reflectivity, Doppler radial velocity, and 
radial-velocity-gradient proxy features within an encoder-decoder U-Net. 
Using the most recent radar volume scan, the model predicts 12 future composite 
reflectivity fields at 7.5-min intervals up to 90 min lead time. The derived velocity 
magnitude, divergence-like, directional-shear, and vorticity-like channels represent 
kinematic signatures associated with convergence and boundary interactions without 
requiring full wind-field retrieval. A high-reflectivity attention module is used to 
improve sensitivity to convective cores, and physics-guided attribution to 
examine whether the learned sensitivities are meteorologically meaningful. 
The model is trained using Mumbai Doppler radar observations from May to August 
2023 and evaluated on temporally independent events. At 90 min lead time, 
Critical Success Index values are 0.437, 0.332, and 0.193 
for $\geq$10, $\geq$20, and $\geq$30 dBZ thresholds, respectively. 
Compared with persistence, the model gives lower RMSE and higher 
spatial correlation at longer lead times. Once trained, it runs on 
a standard computer, generating nowcasts within seconds for real-time use.
\end{abstract}


\section{Introduction}
\IEEEPARstart{A}{ccurate} precipitation prediction is fundamental 
for reducing weather-related risks in cities, particularly for flood 
management, aviation operations, emergency response, and transport planning. 
Within this broader prediction problem, nowcasting focuses on the immediate 
few minutes to hours ahead, typically up to about 0-90 min, when rainfall systems 
can evolve rapidly and decisions must be made with minimal lead time \cite{Wilson1998,Pulkkinen2019}.
This window is especially important for short-duration, high-intensity rainfall spells, which can trigger 
flash flooding in densely built urban catchments within less than an hour. In coastal megacities such as Mumbai, 
rapidly evolving monsoon convection, land-sea interactions, and localized cloudbursts can overwhelm stormwater drainage, 
inundate low-lying roads and underpasses, disrupt transport, and affect aviation operations. 
The extreme rainfall event of 26 July 2005 in Mumbai underscored the need for neighbourhood-scale 
nowcast guidance on where intense convective cells are likely to initiate, intensify, 
and move over the next one to two hours \cite{Mohanty2023}.
In addition, extreme rainfall over Mumbai exhibits strong intra-urban spatial 
variability, with earlier work showing that urban-induced intensification can 
occur in localized pockets and that rainfall correlations between stations may 
become statistically insignificant beyond distances of about 10 km \cite{Paul2018}.
Producing such guidance using radar-assimilated regional numerical 
models remains operationally challenging because frequent data 
assimilation, initialization, and model spin-up must be 
completed within a short decision window. A fast radar-based 
machine-learning framework therefore offers a practical route for generating 
localized nowcasts with the latency required for real-time urban response.

At nowcast horizons, numerical weather prediction (NWP) faces 
structural limitations: kilometre-scale convection is 
highly sensitive to initial-condition errors and model 
spin-up, while microphysics and boundary-layer parameterizations 
introduce biases. In addition, data assimilation and I/O latency 
can exceed decision windows for fast-evolving storms \cite{Bauer2015}.
Even convection-allowing models can improve storm morphology yet still 
exhibit substantial timing and placement errors for convective 
initiation at 0-3 h lead times \cite{Clark2016}.
Weather radar remains the backbone of operational nowcasting 
because it provides high-resolution, near-real-time depictions 
of storm structure and motion. Classical radar-based approaches, 
including Tracking Radar Echoes by Correlation \cite{Rinehart1978} 
and optical-flow extrapolation \cite{Bowler2004}, estimate a motion field and advect 
reflectivity forward under a Lagrangian persistence assumption.
While computationally efficient and skilful at very short lead times, 
these techniques degrade rapidly beyond ~30-60 min because they cannot 
represent convective initiation, growth, or decay \cite{Wilson1998}. 
Probabilistic ensemble schemes such as STEPS merge extrapolation with 
stochastic perturbations and downscaled NWP to quantify uncertainty, 
but they remain constrained by persistence \cite{Bowler2006}.

Open-source libraries such as pysteps have standardized probabilistic 
extrapolation, verification, and benchmarking for radar 
nowcasting \cite{Pulkkinen2019}. 
Recent probabilistic and generative approaches, including DGMR, 
produce sharper ensembles with improved spatial structure 
and categorical skill relative to traditional extrapolation 
methods \cite{Ravuri2021, Augst2017}.

Recent advances in deep learning have shifted radar nowcasting 
from extrapolation-based methods toward data-driven spatiotemporal 
prediction. Early models such as ConvLSTM demonstrated improved 
skill by jointly learning spatial and temporal evolution \cite{Shi2015}, 
while encoder-decoder architectures such as U-Net preserved 
fine-scale morphology through multiscale representations and 
skip connections \cite{Ronneberger2015}. Subsequent developments, including TrajGRU, 
PredRNN, attention-based encoder-forecasters, and large-context 
models such as MetNet and MetNet-2, further improved storm evolution, 
temporal coherence, and forecast skill across extended 
lead times \cite{Shi2017,Wang2017,Sonderby2020,Lam2022}.

More recently, physics-informed nowcasting frameworks have 
incorporated continuity constraints, transport-aware evolution, 
and conservation principles to improve long-lead stability, 
probabilistic calibration, and representation of convective 
evolution. Representative examples include NowcastNet, LUPIN, 
and ClimODE, which integrate physical constraints through 
complementary deterministic, stochastic, or continuous-time 
formulations \cite{Zhang2023,Pavlik2024,Verma2024}. Collectively, 
these studies demonstrate 
the benefits of embedding physical consistency within 
deep-learning nowcasting systems.

Current radar nowcasting frameworks often rely on reflectivity 
alone and miss the kinematic signatures of convergence, shear, 
and boundary interactions that precede convective growth. 
Also, existing deep learning models frequently lack 
scientific interpretability, which limits operational trust. 
To address these gaps, in this study, 
we develop a radar-only deep learning nowcasting framework 
tailored to Mumbai's monsoon convection. We construct multi-elevation, 
multi-variable input tensors that combine reflectivity, 
Doppler radial velocity, and physically motivated proxy 
predictors derived from spatial gradients of radial velocity. 
In contrast to reflectivity-only encoder-decoder baselines, 
this design injects dynamically informative cues into the learned representation 
while avoiding the need for full wind retrieval from single-Doppler measurements. 
The forecasting core is an encoder-decoder U-Net that maps the most recent radar 
volume scan to a 12-step composite-reflectivity forecast sequence at 7.5-min 
intervals (up to 90 min). To improve scientific interpretability and operational trust, 
we additionally apply physics-guided feature attribution to assess whether learned 
sensitivities concentrate on meteorologically meaningful precursor structures, 
such as boundary signatures and convergence-like patterns.

In the present analysis, our goal is to develop a radar-based 
nowcasting framework for Mumbai using a machine learning approach. 
The proposed method is computationally efficient, requires minimal 
processing effort, and can be rapidly updated with each new radar 
scan generated every 7.5 minutes, offering strong potential for 
real-time applications. Building on this framework, we address 
four key objectives: (i) to demonstrate that multi-elevation, multi-variable 
inputs improve 
skill relative to reflectivity-only baselines across standard thresholds 
and lead times; (ii) to evaluate whether radial-velocity-gradient proxy features improve 
representation of convective growth and moderate-to-heavy echoes; (iii) 
to show that attribution maps provide meteorologically consistent explanations 
that are useful for practitioners; and (iv) to verify that the gains are robust across 
temporally disjoint events and feasible within a 7.5-min operational update cycle.

\section{Methods}
This section details the end-to-end methodology underpinning the radar-based 
nowcasting system. We first construct physics-aware inputs by transforming 
multi-elevation Doppler radar measurements into multi-channel tensors, 
including reflectivity, radial velocity, and radial-velocity-gradient kinematic 
proxy fields, and we define multi-step composite-reflectivity targets at fixed 
lead times. We then describe a feed-forward U-Net that maps the most recent 
multi-elevation radar volume scan to a sequence of future reflectivity fields, 
augmented with a high-reflectivity attention mechanism to emphasize convective cores. 
The learning objective combines intensity-weighted reconstruction with structural 
similarity and threshold-aware terms to align training with both spatial realism 
and exceedances skill at operational reflectivity thresholds. We conclude with data 
preparation, channel-wise normalization, and temporally disjoint training/validation/test splits.

\subsection{Radar data \& preprocessing}
Radar data for this study are sourced from the Doppler weather radar operated by the 
India Meteorological Department (IMD) at the Veravali site near Mumbai ($\approx$19.1° N, 72.9° E). 
The C-band volume-scan radar typically performs ~10 elevation sweeps per volume, providing 
reflectivity and Doppler radial-velocity measurements over the Mumbai metropolitan region 
and adjacent coastal/offshore areas. IMD surveillance extends to roughly 400-450 km; however, 
beam broadening, attenuation, and increasing sampling volume degrade effective resolution 
with distance. Accordingly, echoes within ~250 km are retained and mapped to a 
$1\,\mathrm{km} \times 1\,\mathrm{km}$ Cartesian grid.

Preprocessing follows a standard pipeline for converting polar radar volumes into quality-controlled 
Cartesian products suitable for deep learning. Reflectivity and radial-velocity fields undergo quality 
control, including ground-clutter and anomalous-propagation suppression, masking of low signal-to-noise gates, 
and radial-velocity dealiasing using IMD operational procedures. Data at each elevation are then interpolated 
from polar coordinates onto a Cartesian grid using bilinear interpolation at 1 km spacing in both x and y, 
yielding a 501 x 501 domain ($\approx 500\,\mathrm{km} \times 500\,\mathrm{km}$); scans with incomplete azimuthal coverage or severe artefacts 
are discarded \cite{Osrodka2015,Augst2017,Pulkkinen2019}. Reflectivity is 
clipped to a physically plausible range (e.g., -5 to 55 dBZ) to limit outliers, while radial velocity is 
scaled to a fixed interval prior to feature derivation. For model input, each 2-D Cartesian field is 
resampled to $128 \times 128$ to match the network resolution \cite{OchoaRodriguez2019}. Finally, volume 
scans are aligned to a regular 7.5-min temporal grid via linear interpolation in time, and sequences 
with gaps exceeding one scan interval are excluded from training and evaluation.

\subsection{Physics-Aware Feature Engineering}
The forecasting framework employs physics-aware feature engineering to incorporate kinematic 
information from Doppler radar observations. This is motivated by theoretical and observational 
evidence that low-level convergence, shear, and rotation influence convective initiation, 
organization, and storm evolution 
\cite{Weckwerth2006,Mecikalski2006,Markowski2010}. 
For each radar elevation, six input channels are derived: horizontal reflectivity Z, radial velocity $V_r$, 
radial-velocity magnitude $|V_r|$, a divergence-like radial-velocity-gradient proxy $S$, 
a directional shear proxy $\theta$, and a vorticity-like proxy $\omega$.
\begin{equation}
\label{eq:Xe}
X_e = [Z_e, V_{r,e}, |V_{r,e}|, S_e, \theta_e, \omega_e].
\end{equation}
where $X_e$ denotes the multi-channel input tensor for elevation $e$. The radial-velocity magnitude is defined as
\begin{equation}
\label{eq:Vrmag}
\lvert V_{r,e}\rvert = \sqrt{V_{r,e}^2 + V_{\theta,e}^2},
\end{equation}
where $V_{\theta,e}$ is the tangential velocity component at elevation $e$.

The divergence-like radial-velocity-gradient proxy is computed as
\begin{equation}
\label{eq:Se}
S_e = \frac{\partial V_{r,e}}{\partial x} + \frac{\partial V_{r,e}}{\partial y}.
\end{equation}

The directional shear proxy is defined as
\begin{equation}
\label{eq:thetae}
\theta_e = \frac{1}{\pi}\operatorname{atan2}\left(\frac{\partial V_{r,e}}{\partial y}, \frac{\partial V_{r,e}}{\partial x}\right).
\end{equation}
which represents the orientation of the local radial-velocity gradient normalized to [-1,1]. The vorticity-like radial-velocity-gradient proxy is computed as
\begin{equation}
\label{eq:omegae}
\omega_e = \frac{\partial V_{r,e}}{\partial y} - \frac{\partial V_{r,e}}{\partial x}.
\end{equation}

Spatial gradients are estimated on the Cartesian radar grid using central differences:
\begin{equation}
\label{eq:dVdx}
\left.\frac{\partial V_{r,e}}{\partial x}\right|_{i,j} =
\frac{V_{r,e}(i,j+1)-V_{r,e}(i,j-1)}{2\Delta x}.
\end{equation}
\begin{equation}
\label{eq:dVdy}
\left.\frac{\partial V_{r,e}}{\partial y}\right|_{i,j} =
\frac{V_{r,e}(i+1,j)-V_{r,e}(i-1,j)}{2\Delta y}.
\end{equation}
where $\Delta x$ and $\Delta y$ are the Cartesian grid spacing. Where necessary, 
a light smoothing filter is applied to $V_r$ before differencing to reduce noise amplification.

Together, these channels provide a compact representation of reflectivity 
structure and radial-velocity-derived kinematic signatures associated with 
convergence, divergence, shear, and rotation, all of which are relevant to 
convective initiation and storm growth. The six channels are constructed 
independently for each radar elevation and then stacked across $N_elev$ 
elevation angles, giving a total input depth of $6N_{\mathrm{elev}}$. The resulting 
multi-elevation tensor is resampled to a $128 \times 128$ spatial grid before being 
supplied to the network. The prediction targets are column-maximum reflectivity 
fields at 12 future lead times, sampled every 7.5 min and extending to a 
maximum lead time of 90 min.

\subsection{Network Architecture}
The forecasting model is a feed-forward spatio-temporal U-Net, 
designed to predict the full reflectivity evolution sequence 
from a single input time slice. This architecture is chosen for its 
ability to simultaneously handle the spatial complexity of storm 
structures and the temporal evolution of convective systems, which 
is crucial for accurate short-term precipitation nowcasting. The 
encoder-decoder backbone consists of convolution-batch normalization-LeakyReLU 
blocks, with two downsampling stages to capture multi-scale spatial 
features and bilinear upsampling in the decoder to reconstruct 
high-resolution predictions. Skip connections are employed to 
preserve high-frequency storm structures across scales, which 
ensures that fine-grained details (such as the edges of convective cells) 
are maintained throughout the network.
\begin{figure*}[!t]
\centering
\includegraphics[width=\textwidth]{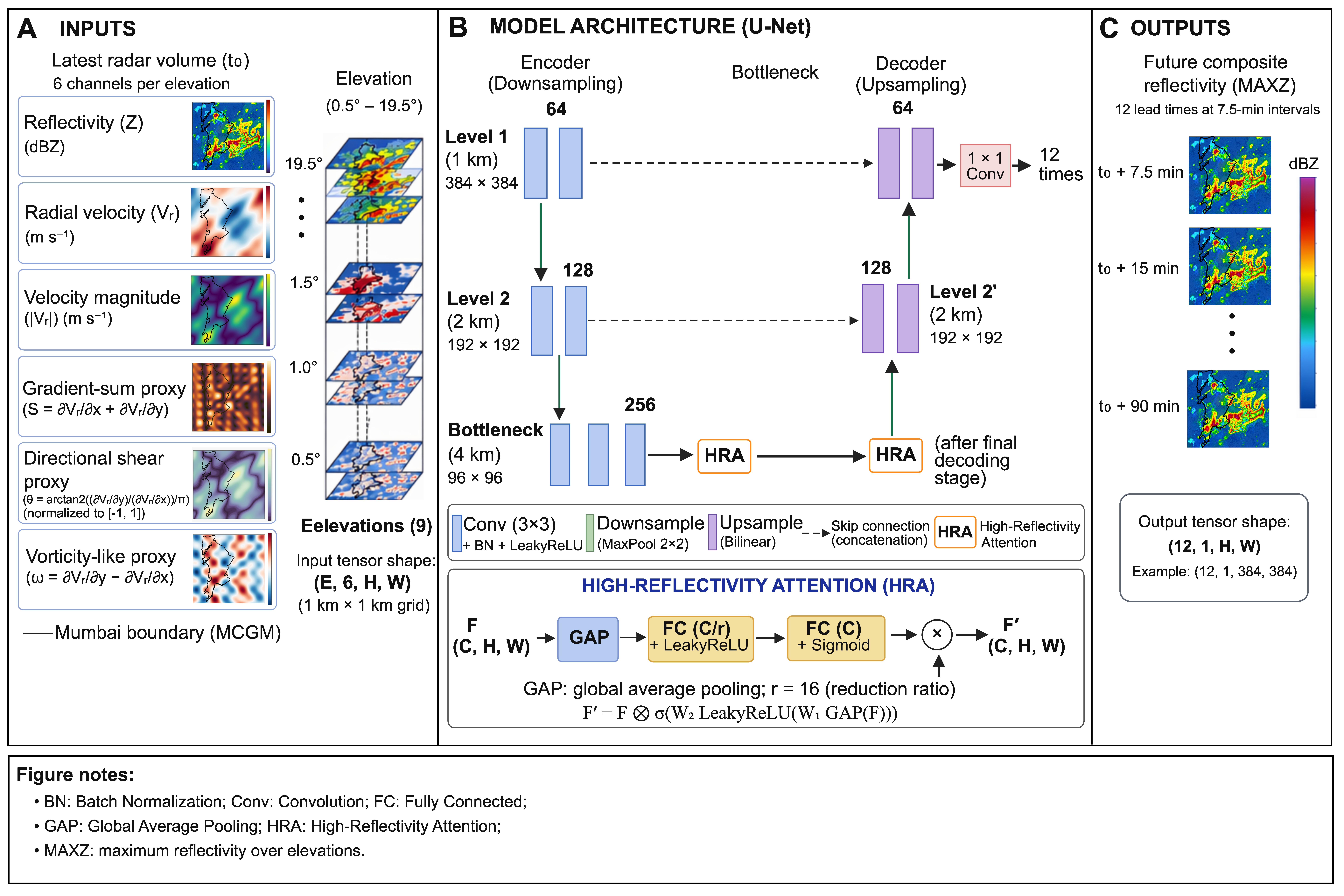}
\caption{U-Net architecture for radar-based weather nowcasting. 
Multi-elevation radar reflectivity and Doppler radial-velocity 
fields from the most recent volume scan (\(\approx 7\) min scan cadence)
are provided as input and augmented through feature engineering 
to derive kinematic descriptors, including velocity magnitude, 
divergence, direction, and vorticity. The network employs a 
two-level encoder-decoder structure with skip connections to 
preserve multiscale spatial information, while a bottleneck 
layer captures compact representations. During decoding, bilinear 
upsampling and feature concatenation progressively restore spatial 
resolution. A high-reflectivity attention module is applied after 
the final decoding stage to emphasize regions associated with 
intense precipitation. The model outputs T future reflectivity 
frames corresponding to a 90-min forecast horizon at approximately 
scan-time (\(\approx 7\) min) intervals.}
\label{fig_1}
\end{figure*}

We developed a U-Net-based encoder-decoder model to forecast short-term 
radar reflectivity. The network ingests the most recent \(128 \times 128\) gridded 
fields at time \(t_0\) of radar reflectivity (dBZ) and Doppler radial 
velocity (\(\mathrm{m\,s^{-1}}\)) sampled at the native radar scan cadence 
(\(\approx 7\) min). To enrich the dynamical information available to the 
model, additional kinematic predictors are computed from the radial-velocity 
fields and concatenated with the raw inputs prior to encoding, including 
velocity magnitude \(\lvert v \rvert\), a divergence-like gradient proxy 
\(\partial v / \partial y\), and a direction proxy derived from the gradient orientation
defined by \(\operatorname{atan2}(\partial v/\partial y,\partial v/\partial x)/\pi\).

The encoder comprises two convolutional stages with \(3 \times 3\) kernels and increasing 
channel depth (64 and 128). Each stage is followed by batch normalization and 
LeakyReLU activation, and max pooling is used for spatial downsampling. 
A 256-channel bottleneck layer captures compact, high-level representations 
of the input fields. The decoder reconstructs full resolution using bilinear 
upsampling followed by \(3 \times 3\) convolutions, and integrates skip connections 
from the corresponding encoder stages to preserve fine-scale spatial structure. 
A final \(1 \times 1\) convolution produces the forecast sequence, yielding 12 future 
reflectivity frames (\(\approx 90\) min lead time at scan-time intervals).

To emphasize intense precipitation regions that are most relevant for 
convective nowcasting, we incorporate a high-reflectivity attention (HRA) 
module at the bottleneck and after the final decoding stage. Given a 
reflectivity feature representation, the attention map $A$ is computed using 
global spatial pooling followed by a two-layer gating function:

\begin{equation}
\label{eq:A}
A = \sigma\left(W_2\,\mathrm{ReLU}(W_1\bar{Z})\right).
\end{equation}
where $\bar{Z}$ is the spatially averaged reflectivity feature map, $W_1$ and $W_2$ 
are trainable parameters implemented as fully connected layers or 
equivalently $1 \times 1$ convolutions, and $\sigma$ denotes the sigmoid activation. 
The attended feature tensor is then obtained through residual reweighting:

\begin{equation}
\label{eq:F}
\hat{F} = F \odot (1 + \alpha A)
\end{equation}
where F is the input feature tensor, 
$\odot$ denotes element-wise multiplication, and $\alpha$controls the 
strength of attention-based amplification. In this study, $\alpha$ is 
set to 1.2. By selectively amplifying features associated with high 
reflectivity, the HRA mechanism increases the network's 
sensitivity to convective cores and improves the reconstruction 
and prediction of rapidly evolving storm structures.

\subsection{Loss Functions}
Training minimizes a composite objective that balances amplitude fidelity, 
spatial structure, and operational skill at precipitation thresholds:

\begin{equation}
\label{eq:total_loss}
L =
\lambda_{\mathrm{iw}} L_{\mathrm{iwMSE}}
+
\lambda_{\mathrm{SSIM}} (1-\mathrm{SSIM})
+
\lambda_{\mathrm{thr}} L_{\mathrm{thr}}.
\end{equation}

where SSIM denotes the Structural Similarity Index Measure, which quantifies 
the similarity between predicted and observed reflectivity fields in terms 
of luminance, contrast, and structural consistency. This formulation combines 
an amplitude-focused reconstruction term, a structural similarity term, and 
a threshold-aware term, so that training is aligned with both pixel-wise 
accuracy and operational verification metrics. The weighting coefficients 
\(\lambda_{\mathrm{iw}}\), \(\lambda_{\mathrm{SSIM}}\), and 
\(\lambda_{\mathrm{thr}}\) are chosen such that no single component dominates 
the gradient, yielding forecasts that retain realistic storm morphology while 
maintaining skill at application-relevant reflectivity thresholds. Through the 
reflectivity-dependent weighting in \(L_{\mathrm{iwMSE}}\), the loss places 
larger penalties on errors in moderate and high reflectivity regions 
(\(\geq 25\)--35 dBZ), so the optimizer concentrates capacity on convective 
cells and heavy rain rather than weak, low-impact echoes.

To emphasize convective echoes, the reconstruction error is weighted by 
reflectivity in dBZ. The weights are computed in physical reflectivity space, 
while the errors are evaluated in the model's normalized space for numerical 
stability. Let \(Z_t(x,y)\) and \(\hat{Z}_t(x,y)\) denote the observed and 
predicted reflectivity fields at lead time \(t = 1,\ldots,T\). The 
reflectivity-aware weight is defined as

\begin{equation}
\label{eq:weight_function}
w(Z)=
\begin{cases}
\alpha_0, & Z < 15~\mathrm{dBZ}, \\
\alpha_1, & 15 \leq Z < 25~\mathrm{dBZ}, \\
\alpha_2, & 25 \leq Z < 35~\mathrm{dBZ}, \\
\alpha_3, & Z \geq 35~\mathrm{dBZ}
\end{cases}
\end{equation}

with
\[
\alpha = \{1.0,\,1.3,\,1.8,\,2.5\}.
\]

The intensity-weighted mean squared error is then computed over all forecast 
lead times and spatial grid points as

\begin{equation}
\label{eq:iwmse}
\begin{aligned}
L_{\mathrm{iwMSE}}
&=
\frac{1}{THW}
\sum_{t=1}^{T}
\sum_{x=1}^{W}
\sum_{y=1}^{H}
w\!\left(Z_t^{\mathrm{dBZ}}(x,y)\right)
\\
&\quad \times
\left(
\hat{Z}_t^{\mathrm{norm}}(x,y)
-
Z_t^{\mathrm{norm}}(x,y)
\right)^2.
\end{aligned}
\end{equation}

where ``dBZ'' and ``norm'' denote the physical and normalized domains, 
respectively. The affine normalization is inverted only for computing the 
reflectivity-dependent weight \(w(\cdot)\).

\subsection{Training Configuration}
Training data are drawn from Doppler radar observations collected over 
Mumbai between May and August 2023, with an 80--10--10 split for 
training, validation, and testing. Input variables are standardized 
by channel to ensure consistent scaling across the dataset. During 
training, data augmentation techniques such as random cropping and 
horizontal flipping are applied to increase model robustness to 
spatial variability. The model is trained using the Adam optimizer 
(learning rate \(1 \times 10^{-4}\)) with early stopping based on 
validation SSIM to prevent overfitting. Additionally, a learning rate 
scheduler is used to adjust the learning rate during training to 
enhance convergence and optimize training efficiency. The model runs on an 
NVIDIA GPU (CUDA-enabled) with a batch size of 8 
and generates reflectivity predictions every 7.5 min up to 90 min 
ahead. Validation performance is continuously monitored using metrics 
such as SSIM and the Critical Success Index (CSI), ensuring that the 
model not only fits well but also generalizes effectively. Test data, 
withheld from training and validation, are used to assess the model's 
ability to generalize to unseen radar scans and to evaluate 
performance on a real-world test set.

\subsection{Evaluation metrics}
Model performance was assessed using both categorical and continuous 
verification metrics. Categorical skill was evaluated at reflectivity 
thresholds of \(\geq 10\), \(\geq 20\), and \(\geq 30\) dBZ using the 
Critical Success Index (CSI), Equitable Threat Score (ETS), frequency 
bias, Probability of Detection (POD), False Alarm Ratio (FAR), and 
False Positive Rate (FPR). Let \(H\), \(M\), \(F\), and \(CN\) denote 
hits, misses, false alarms, and correct negatives, respectively. 
These metrics are defined as

\begin{equation}
\label{eq:csi}
\mathrm{CSI} = \frac{H}{H+M+F}
\end{equation}

\begin{equation}
\label{eq:ets}
\mathrm{ETS} =
\frac{H-H_r}{H+M+F-H_r}
\end{equation}

\begin{equation}
\label{eq:random_hits}
H_r =
\frac{(H+M)(H+F)}
{H+M+F+CN}
\end{equation}

\begin{equation}
\label{eq:bias}
\mathrm{Bias} =
\frac{H+F}{H+M}
\end{equation}

\begin{equation}
\label{eq:pod}
\mathrm{POD} =
\frac{H}{H+M}
\end{equation}

\begin{equation}
\label{eq:far}
\mathrm{FAR} =
\frac{F}{H+F}
\end{equation}

\begin{equation}
\label{eq:fpr}
\mathrm{FPR} =
\frac{F}{F+CN}
\end{equation}

Continuous performance was evaluated using the domain-averaged 
root-mean-square error (RMSE) and spatial correlation of composite 
reflectivity:

\begin{equation}
\label{eq:rmse}
\mathrm{RMSE}
=
\sqrt{
\frac{1}{N}
\sum_{i=1}^{N}
(P_i-O_i)^2
}
\end{equation}

\begin{equation}
\label{eq:correlation}
\begin{aligned}
r
&=
\frac{
\sum_{i=1}^{N}
(P_i-\bar{P})(O_i-\bar{O})
}
{
\sqrt{
\sum_{i=1}^{N}
(P_i-\bar{P})^2
}
\sqrt{
\sum_{i=1}^{N}
(O_i-\bar{O})^2
}
}
\end{aligned}
\end{equation}

where \(P_i\) and \(O_i\) denote predicted and observed reflectivity 
values at grid point \(i\), respectively; \(\bar{P}\) and \(\bar{O}\) 
are their spatial means; and \(N\) is the total number of grid cells. 
To assess scale-dependent structural realism, radially averaged power 
spectral density (PSD) was compared between forecasts and observations 
at selected lead times. Observed event support was also reported to 
provide context on threshold-dependent sample prevalence.

\section{Results}
This section reports the evaluation of the proposed radar nowcasting framework 
on an independent test set. We summarize performance across reflectivity thresholds 
using categorical and continuous verification metrics, examine how skill changes 
with lead time relative to a persistence baseline, and assess scale dependent 
structure using spectral and neighbourhood verification. Representative case 
diagnostics are also provided to illustrate dominant error modes.

\subsection{Overall nowcasting performance and intensity dependence}
The spatiotemporal U-Net was evaluated on an independent test dataset using 
standard categorical and continuous verification metrics across reflectivity 
thresholds of $\geq$ 10, $\geq$ 20, and $\geq$ 30 dBZ. Unless otherwise noted, reported values 
are aggregated across all 12 forecast steps (7.5-90 min). The model shows 
strongest skill for weaker and moderate echoes, with performance gradually 
decreasing as the reflectivity threshold increases, consistent with the greater 
difficulty of predicting compact and rapidly evolving convective cores. Across 
thresholds, the model maintains useful categorical skill while false-alarm 
behaviour remains controlled, indicating good discrimination between precipitating 
and non-precipitating regions despite pronounced class imbalance. Overall, the 
results indicate robust performance for weak-to-moderate precipitation, 
with reduced skill for the most intense echoes.

\begin{figure*}[!t]
\centering
\includegraphics[width=\textwidth]{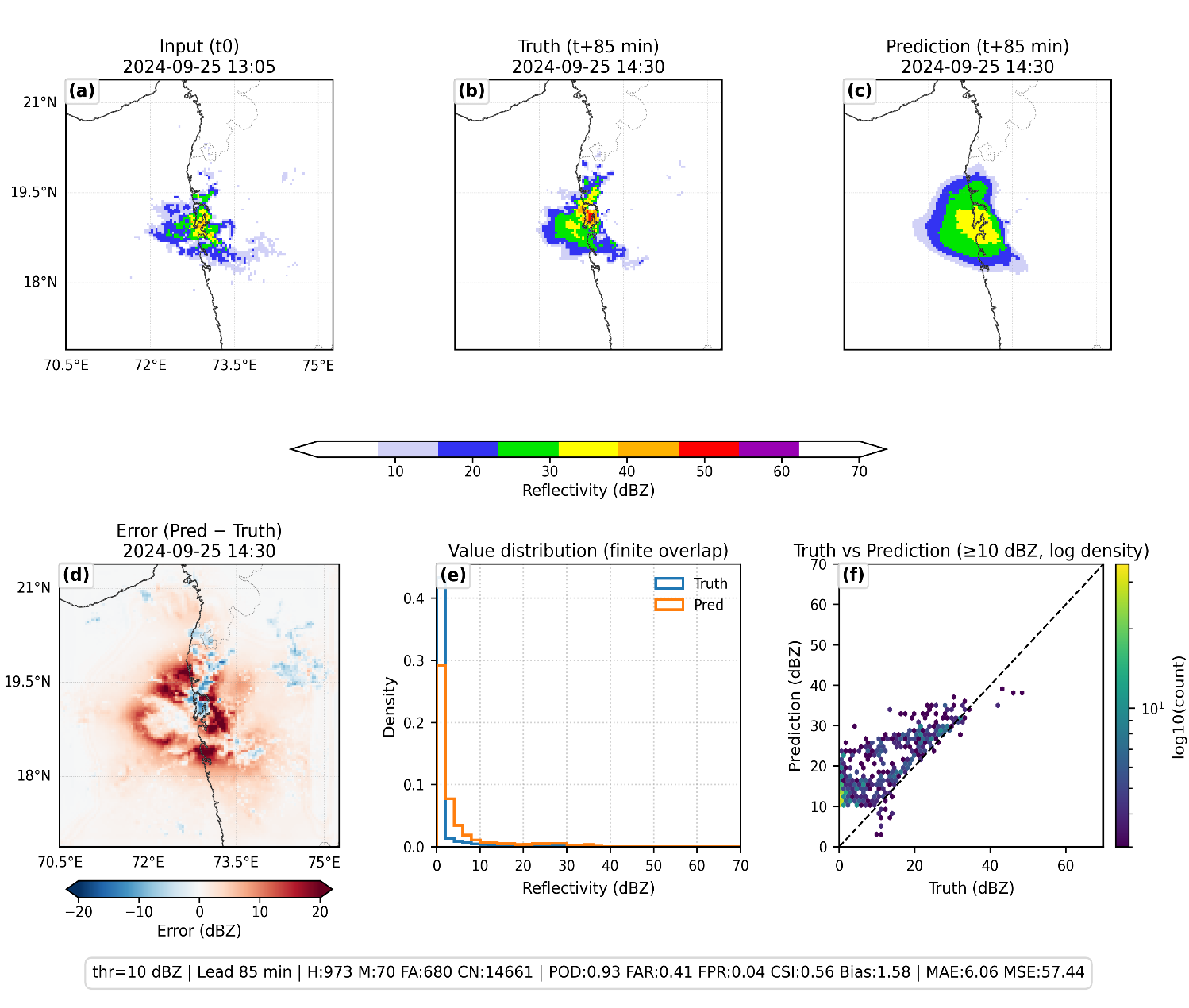}
\caption{Radar nowcasting verification diagnostics for an extreme 
precipitation event on 25 September 2024 
(lead \(= 85\) min). This specific high-impact case was selected 
as a stress test for the model; on this day, rapid convective 
intensification delivered over 100 mm of rainfall in a few hours, 
triggering an IMD Red Alert and severe urban flooding in Mumbai. 
Composite 
reflectivity fields, \(Z\) (dBZ), are shown over the study domain 
with coastline overlay and latitude/longitude axes. The top row shows 
(a) the last input frame at \(t_0\), (b) the verifying observation at 
\(t_0 + 85\) min, and (c) the model prediction for the same valid 
time; all panels use an identical reflectivity color scale. The bottom 
row presents (d) the spatial error field, 
\(\Delta Z = Z_{\mathrm{pred}} - Z_{\mathrm{truth}}\) (dBZ), where 
positive (negative) values indicate overprediction (underprediction); 
(e) marginal reflectivity distributions over finite-overlap pixels for 
the truth and prediction; and (f) the pixel-wise truth--prediction 
joint density (hexbin; log-scaled counts) for pixels with truth 
\(\geq 10\) dBZ, with the 1:1 reference line. The footer reports 
categorical verification at the 10 dBZ exceedance threshold (hits, 
misses, false alarms, and correct negatives) and derived scores 
(POD, FAR, FPR, CSI, and frequency bias), together with continuous 
intensity errors (MAE and MSE).}
\label{fig_2}
\end{figure*}

\subsection{Spectral characteristics (PSD) and loss of small-scale power}
To assess whether forecasts preserve spatial structure across spatial scales, 
we compare the radially averaged power spectral density (PSD) of the predicted 
fields against the verifying observations at selected lead times shown in Figure 4, 
spanning approximately 8, 31, 62, and 85 min. At ~8 min, the model spectrum is 
close to the observed spectrum at larger wavelengths, while showing reduced 
power at shorter wavelengths, indicative of smoothing and attenuation of fine-scale 
variability. With increasing lead time ($\approx$31, $\approx$62, and $\approx$85 min), 
the deficit in short-wavelength power becomes more pronounced, consistent 
with progressive weakening of sharp gradients and compact convective features. 
The optical-flow baseline generally retains more short-wavelength power than the model, 
although both methods diverge increasingly from the observed spectrum as lead time increases. 
Overall, the PSD diagnostics complement the categorical metrics by indicating that 
forecast degradation is associated primarily with loss of small-scale structure and 
intensity damping, rather than the introduction of spurious variability.

\begin{figure*}[!t]
\centering
\includegraphics[width=0.94\textwidth]{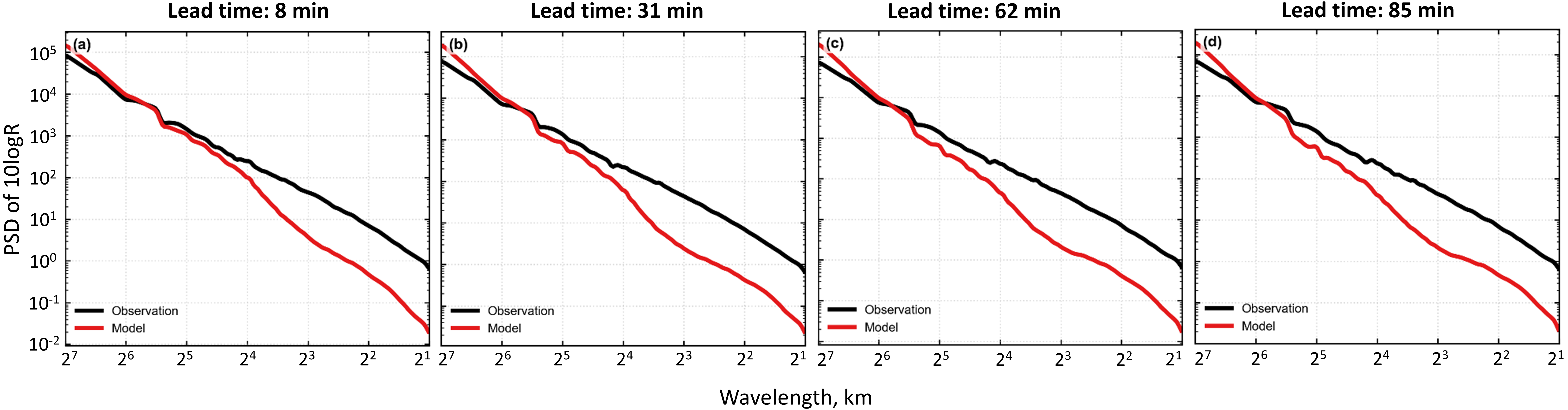}
\caption{Radially averaged power spectral density (PSD) 
of composite reflectivity (10logR) as a function of wavelength 
for the observations, optical-flow extrapolation, and the 
proposed model at selected lead times of approximately 8, 31, 62, and 85 min.}
\label{fig_3}
\end{figure*}

\subsection{Lead time dependence of categorical skill}
To further examine the temporal robustness of the proposed framework, 
we next evaluate how categorical skill varies with forecast lead time 
across weak, moderate, and intense reflectivity thresholds. Figure~5 
shows that the proposed model maintains useful categorical skill 
throughout the 90-min forecast horizon, with 90-min CSI values of 
0.437, 0.332, and 0.193 for the \(\geq 10\), \(\geq 20\), and 
\(\geq 30\) dBZ thresholds, respectively. Forecast skill decreases 
gradually with increasing lead time, indicating stable temporal 
behaviour rather than an abrupt loss of predictability. For 
\(\geq 20\) dBZ, CSI declines from 0.501 at 7.5 min to 0.387 at 
60 min, and to 0.332 at 90 min. ETS (Equitable Threat Score, which 
corrects CSI for random hits) follows the same pattern, decreasing 
from 0.484 at 7.5 min to 0.367 at 60 min and 0.314 at 90 min. 
Frequency bias for \(\geq 20\) dBZ remains above unity at all lead 
times (for example, 1.168 at 7.5 min and 1.228 at 90 min), indicating 
persistent overprediction of areal coverage for moderate echoes even 
as CSI decreases.

For \(\geq 10\) dBZ, CSI decreases from 0.596 at 7.5 min to 0.496 
at 60 min, and to 0.437 at 90 min, with bias remaining close to 
unity (approximately 1.07--1.10), indicating only slight 
overcoverage. In contrast, intense echoes (\(\geq 30\) dBZ) show 
the strongest degradation and clear underprediction at long lead 
times. CSI decreases from 0.423 at 7.5 min to 0.283 at 60 min, and 
to 0.193 at 90 min. Frequency bias drops from 0.723 at 7.5 min to 
0.492 at 90 min. This confirms that the primary limitation at 
extended lead times is the representation of compact high-reflectivity 
cores.

\begin{figure*}[!t]
\centering
\includegraphics[width=\linewidth]{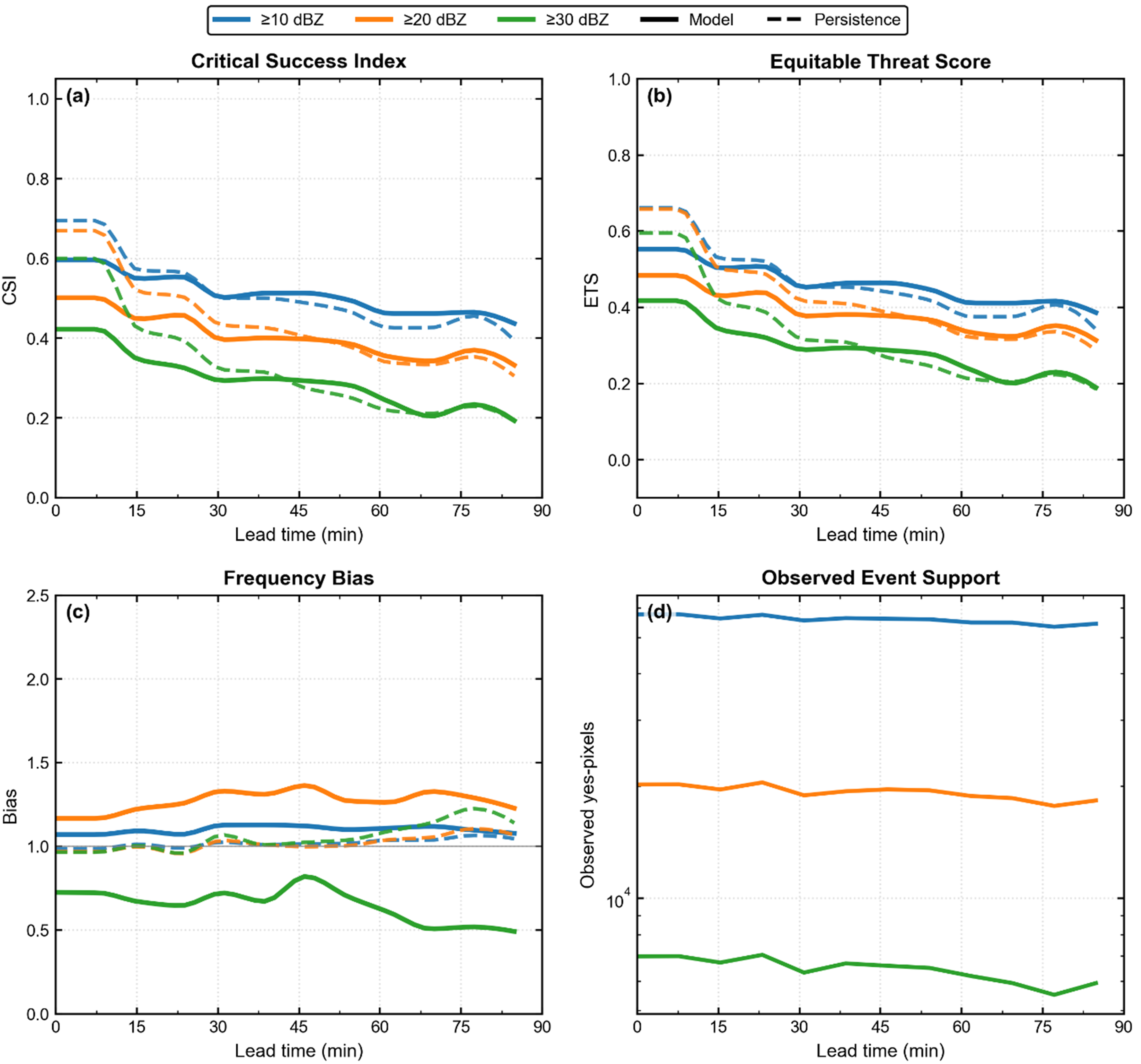}
\caption{Lead-time evolution of categorical verification 
metrics for the proposed model and a persistence baseline 
across reflectivity thresholds of $\geq 10$, $\geq 20$, and $\geq 30$ dBZ. 
Panels show (a) Critical Success Index (CSI), (b) Equitable 
Threat Score (ETS), (c) frequency bias, and (d) observed 
event support as functions of forecast lead time. Solid 
lines denote the proposed model, and dashed lines denote 
persistence. The model retains useful skill across the 
90-min forecast horizon, with the highest performance at 
lower reflectivity thresholds and increasing 
underprediction for intense echoes at longer lead times.}
\label{fig_4}
\end{figure*}

\subsection{Continuous verification against a persistence baseline}
We next examine lead time dependence using continuous verification 
against a persistence baseline (Fig. 5), based on domain-averaged 
RMSE and spatial correlation of composite reflectivity. Persistence 
is most competitive at the shortest horizon: at 7.5 min, it achieves 
RMSE 2.315 dBZ and correlation 0.897, compared with 3.411 dBZ and 
0.846 for the model. This short-lead advantage of persistence is 
expected and consistent with the literature, since no learned model 
can improve upon the trivial first-step extrapolation before storm 
structures have had time to evolve. By approximately 31 min, the 
model becomes competitive with persistence, with slightly lower 
RMSE (4.275 dBZ versus 4.310 dBZ) and higher spatial correlation 
(0.774 versus 0.748). Beyond roughly 35 to 40 min, persistence 
degrades more rapidly. At approximately 54 min, the model shows 
lower RMSE (4.416 dBZ versus 4.801 dBZ) and higher correlation 
(0.768 versus 0.712), and by 85 min the separation increases 
further, with model RMSE 4.746 dBZ and correlation 0.714 relative 
to persistence RMSE 5.461 dBZ and correlation 0.629. Overall, 
the model maintains stronger spatial agreement and lower error 
at longer horizons, indicating improved representation of 
evolving storm structure relative to pure persistence.

\begin{figure*}[!t]
\centering
\includegraphics[width=\linewidth]{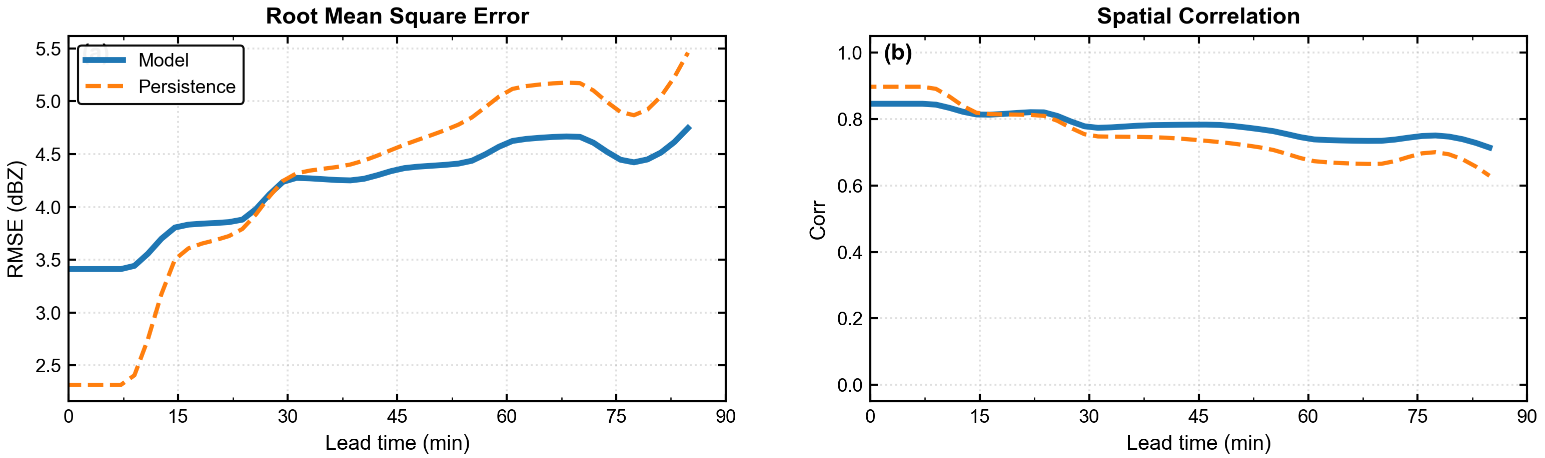}
\caption{Lead-time evolution of continuous verification 
metrics for composite reflectivity, comparing the proposed 
model (solid) with a persistence baseline (dashed): (a) 
domain-averaged RMSE (dBZ) and (b) spatial correlation 
between forecast and observation. Persistence is most 
competitive at the shortest lead times, while the proposed 
model retains lower RMSE and higher spatial 
correlation at longer horizons.}
\label{fig_5}
\end{figure*}

\subsection{Station-Wise Multi-Event Validation}
To further evaluate operational applicability, 
the proposed framework was additionally validated 
against rainfall observations from 35 automatic weather 
stations across the Mumbai metropolitan region during 
three independent high-impact rainfall 
events (17--19 August 2025). The model reproduced the 
observed spatial distribution of moderate-to-heavy rainfall 
and showed a gradual reduction in station-wise agreement 
with increasing lead time. Detailed station-wise comparisons 
and lead-time correlation analyses are provided in the 
Supplementary Material (Section~S1).

Overall, the station-wise validation across multiple extreme-rainfall 
days demonstrates that the proposed framework effectively captures 
moderate-to-heavy rainfall variability at the station scale. The model 
shows consistent performance across independent high-impact events, 
with strong agreement in moderate-rainfall regimes and controlled 
underestimation in extreme cases. These results reinforce the model's 
generalization capability and highlight its suitability for urban-scale 
operational nowcasting, where accurate spatial representation of 
rainfall distribution is critical. In addition, the lead-time correlation 
analysis shows that station-wise agreement degrades 
gradually across successive 15-minute windows rather than collapsing 
abruptly, indicating stable temporal behaviour across all three validation days.

\section{Discussion}
The results indicate that fusing multi-elevation radar observations 
with Doppler radial velocity--derived kinematic proxy channels within 
a U-Net-style encoder--decoder can yield robust short-range 
precipitation nowcasts in a monsoon-dominated coastal environment. 
Traditional extrapolation methods are effective at very short lead 
times, but skill often degrades beyond \(\sim 30\)--60 min because 
Lagrangian persistence cannot represent convective initiation, 
intensification, or decay \cite{Rinehart1978, Wilson1998, Bowler2004}. Deep learning nowcasting models can 
partially address this limitation by learning nonlinear structure and 
evolution from radar archives, as demonstrated by ConvLSTM sequence 
learners and encoder--decoder families such as U-Net \cite{Shi2015, Ronneberger2015}. 
Consistent with prior work, our 
evaluation shows gradual skill degradation with lead time, controlled 
false-alarm behaviour, and near-neutral frequency bias for 
light-to-moderate echoes. These characteristics suggest that the model 
captures storm-scale organization and displacement while limiting 
spurious precipitation, a behaviour commonly reported for deterministic 
radar nowcasts that preserve coherent morphology even as intensity 
errors grow with forecast horizon \cite{Sonderby2020, Lam2022}.

A pronounced intensity dependence remains evident. Skill decreases at 
higher reflectivity thresholds, and intense convective cores are 
underpredicted at longer lead times. Such attenuation and smoothing of 
sharp reflectivity gradients are a widely recognized limitation of 
deterministic point prediction in radar nowcasting, particularly for 
rare, compact extremes that evolve rapidly through nonlinear growth 
and decay processes \cite{Wilson1998, Ravuri2021, Sonderby2020, Lam2022}. Importantly, the dominant 
error mode in the present study is weakening or missing of intense 
echoes rather than widespread false echoes, as reflected by 
persistently low false positive rates and high specificity across 
thresholds. This conservative error behaviour is operationally 
meaningful in applications such as urban flood early warning and 
aviation support, where false alarms can be costly and where reliable 
spatial guidance is often preferred over aggressive intensity 
amplification.

In addition to categorical verification, the spectral diagnostics 
provide an objective view of scale-dependent realism. Fig. ~3 shows 
that the radially averaged power spectral density progressively 
under-represents short-wavelength variance with increasing lead time, 
indicating smoothing and loss of fine-scale structure, while 
larger-scale variability is comparatively better retained. This 
behaviour is consistent with a known limitation of deterministic 
encoder--decoder nowcasting, where optimization with pixel-wise 
objectives tends to damp sharp gradients and compact convective 
maxima. The PSD result therefore supports the interpretation that the 
dominant error mode is intensity attenuation and boundary smoothing 
rather than widespread generation of spurious echoes, and it motivates 
future extensions that better preserve small-scale variability 
(e.g., scale-aware losses, multi-scale supervision, or 
probabilistic/generative formulations).

Within the broader methodological landscape, the proposed 
framework occupies a complementary position. Large-context models 
such as MetNet and MetNet-2 improve regional consistency through 
multiscale context aggregation \cite{Sonderby2020,Lam2022}. 
Generative and physics-constrained approaches, including DGMR, 
NowcastNet, LUPIN, and ClimODE, focus on probabilistic 
forecasting and physically constrained evolution to enhance 
uncertainty representation, calibration, and long-lead 
forecast skill \cite{Ravuri2021,Zhang2023,Pavlik2024,Verma2024}.
In contrast, the present study emphasizes a compact, deterministic, radar-only 
architecture that incorporates radial-velocity-gradient proxy 
channels without requiring full wind retrieval. The ability to 
preserve mesoscale structure, orientation, and coherent 
displacement at extended lead times suggests that these proxies 
regularize the learned evolution during organized monsoon 
convection, consistent with evidence that convergence- and 
shear-related signatures contain useful information for 
convective initiation and storm 
evolution \cite{Weckwerth2006, Mecikalski2006, Mecikalski2015, Zeng2019}. 
A further objective of this study is to improve operational 
trust through interpretability. The attribution analysis 
links model sensitivity to recognizable storm-scale features, 
including convergence-line signatures and gust-front-like 
boundaries, and verifies that the network learns physically 
meaningful cues rather than artefacts, consistent with recent 
work in interpretable machine learning for weather 
and climate applications \cite{EbertUphoff2020, Yang2024}. 
Case-based diagnostics (Fig. ~2) further show that the model 
reproduces the location, orientation, and spatial extent of 
organized systems, with errors primarily arising from 
intensity damping and modest displacement rather than widespread spurious precipitation.

Despite these strengths, underprediction of intense cores remains the 
primary limitation and motivates clear directions for improvement. 
Promising pathways include hybrid deterministic--stochastic 
formulations, explicit evolution constraints, and training objectives 
that better emphasize rare extremes while preserving the conservative 
false-alarm behaviour observed here \cite{Ravuri2021, Zhang2023}. 
Incorporating the proposed kinematic proxy representation 
into probabilistic or generative frameworks may be particularly 
beneficial, as it offers a mechanism to retain the spatial reliability 
and dynamical structure captured by the deterministic model while 
improving representation of high-intensity convection.

\section{Conclusions}
This study presents a compact, radar-only deep learning framework 
for short-range precipitation nowcasting tailored to Mumbai's 
monsoon convection. The approach fuses multi-elevation reflectivity 
with Doppler radial velocity and radial-velocity--gradient proxy 
channels within a spatiotemporal U-Net, and produces 12-step 
composite-reflectivity forecasts at 7.5 min intervals up to 
approximately 90 min lead time. The proxy representation is designed 
to encode convergence- and shear-related kinematic signatures from 
single-Doppler observations without requiring retrieval of a full 
two-dimensional wind field, while a high-reflectivity attention 
mechanism improves sensitivity to convective cores.

Evaluation on temporally disjoint events shows that forecast skill 
degrades gradually with lead time, while false alarms remain 
controlled across thresholds, indicating stable behaviour beyond the 
\(\sim 30\)--60 min range where extrapolation-based methods typically 
deteriorate. The model retains useful categorical skill throughout 
the forecast horizon, with 90-min CSI values of 0.437, 0.332, and 
0.193 at \(\geq 10\), \(\geq 20\), and \(\geq 30\) dBZ, respectively. 
Performance remains intensity dependent, with the strongest skill at 
lower thresholds and systematic underprediction of compact 
high-reflectivity cores at longer lead times. Continuous verification 
further shows that although persistence is strongest at the shortest 
lead time, the proposed model achieves lower RMSE and higher spatial 
correlation at longer horizons, indicating improved representation of 
evolving storm structure.

Physics-guided attribution indicates that the network's sensitivity 
is concentrated in physically interpretable regions, highlighting 
kinematic precursor patterns near developing cells and along 
boundary-like features consistent with convergence lines and gust 
fronts. Together, these findings support the value of physics-aware, 
multi-variable radar inputs for improving deterministic nowcast 
robustness and interpretability in a tropical coastal setting.

Future work should focus on improving extreme-core prediction through 
probabilistic or hybrid deterministic--stochastic formulations, 
incorporating explicit evolution or continuity constraints, and 
refining objective functions and sampling strategies to better 
represent rare, high-intensity convection while preserving the 
conservative false-alarm characteristics demonstrated here.

\section*{Acknowledgments}
This project has received funding and support from the HDFC 
ERGO - IIT Bombay Innovation Lab, a partnership between HDFC ERGO 
General Insurance Company Ltd. and Indian Institute of Technology Bombay 
committed to support innovation and entrepreneurship in insurance and related areas.
\bibliographystyle{IEEEtran}
\bibliography{references}

\vfill

\end{document}